\documentclass{article}
\usepackage{ijcai22}
\usepackage{times}

\usepackage{soul}
\usepackage{url}
\usepackage[hidelinks]{hyperref}
\usepackage[utf8]{inputenc}
\usepackage[small]{caption}
\usepackage{graphicx}
\usepackage{amsmath}
\usepackage{booktabs}
\usepackage{longtable}
\usepackage{mydef}
\title{A Survey of Pre-training on Graphs: Taxonomy, Methods and Applications}
\author{
Jun Xia{\rm\textsuperscript{1,2}}\and
Yanqiao Zhu{\rm\textsuperscript{3,4}}\and
Yuanqi Du{\rm\textsuperscript{5}}\And
Stan Z. Li{\rm\textsuperscript{1,2}}\thanks{Corresponding author}\\
\affiliations
\textsuperscript{1}School of Engineering, Westlake University\\
\textsuperscript{2}Institute of Advanced Technology, Westlake Institute for Advanced Study\\
\textsuperscript{3}Center for Research on Intelligent Perception and Computing, \\Institute of Automation, Chinese Academy of Sciences\\
\textsuperscript{4}School of Artificial Intelligence, University of Chinese Academy of Sciences\\
\textsuperscript{5}Department of Computer Science, George Mason University
\emails
\{xiajun, stan.zq.li\}@westlake.edu.cn, yanqiao.zhu@cripac.ia.ac.cn, ydu6@gmu.edu
}

\begin{document}

\maketitle

\begin{abstract}
Pre-trained Language Models (PLMs) such as BERT have revolutionized the landscape of natural language processing (NLP). Inspired by their proliferation, tremendous efforts have been devoted to pre-trained graph models (PGMs) recently. Owing to its huge model parameters, PGMs can capture abundant knowledge from massive labeled and unlabeled graph data. The knowledge implicitly encoded in model parameters can benefit various downstream tasks and help to alleviate several fundamental issues of learning on graphs. In this paper, we provide a comprehensive survey for PGMs. We first briefly present the limitations of graph representation learning and thus introduce the motivation for graph pre-training. Next, we systematically categorize existing PGMs based on a taxonomy from five different perspectives including the history, model architectures, pre-training strategies, tunning strategies and applications. Finally, we outline several promising research directions that can serve as a guideline for future studies.
\end{abstract}

\section{Backgrounds}
The developments of Deep Neural Networks (DNNs) have revolutionized many machine learning tasks in recent years, ranging from image recognition to natural language processing. However, there are still many non-Euclidean graph datasets in real-world applications such as social networks and biochemical graphs which existing neural networks can not handle with. Recent years have witnessed the prosperity of Graph Neural Networks (GNNs)~\cite{wu2020comprehensive} that extend deep learning approaches for such graph-structured data. However, two fundamental challenges impede the wider usage of existing supervised learning on graph datasets:
(1)\emph{Scarce Labeled Data:} Task-specific labeled data can be extremely scarce especially for biochemical domains where high-quality data labeling often requires time-consuming and resource-costly wet-lab experiments. 
(2) \emph{Out-of-distribution Generalization:} Existing GNNs lack out-of-distribution generalization abilities so that their performance substantially degrades when there exist distribution shifts between training and testing graph data. Indeed, nearly all of the deep learning domains are confronted with these challenges. To overcome these challenges, certain progress has been made. For example, the paradigm of PLMs is thriving in NLP community. Specifically, they first pre-train the models on large-scale corpus and then fine-tune these models in various downstream tasks. It is widely recognized that this paradigm can provide a better initial point across downstream tasks and leads to wider optima with better generalization than training from scratch~\cite{hao2019visualizing}. With the emergence of Transformer architecture~\cite{vaswani2017attention}, PLMs such as BERT~\cite{devlin2019bert} have emerged as an dominative role for NLP, which have established state-of-the-arts results for a large variety of NLP tasks.

Inspired by the proliferation of PLMs, tremendous efforts have been devoted to pre-trained graph models (PGMs) recently. In this paper, we present a survey to provide researchers a synthesis and pointer to related research on PGMs. Existing surveys related to this area have only partially focused on self-supervised learning on graphs~\cite{liu2021graph,xie2021self}, but did not go broader to the other important ingredients of PGMs such as supervised pre-training, tunning strategies, various extensions, their applications and etc.
Overall, the contributions can be summarized as follows:

\tikzstyle{leaf}=[draw=hiddendraw,
    rounded corners,minimum height=1.2em,
    fill=hidden-orange!40,text opacity=1, align=center,
    fill opacity=.5,  text=black,align=left,font=\scriptsize,
inner xsep=3pt,
inner ysep=1pt,
]

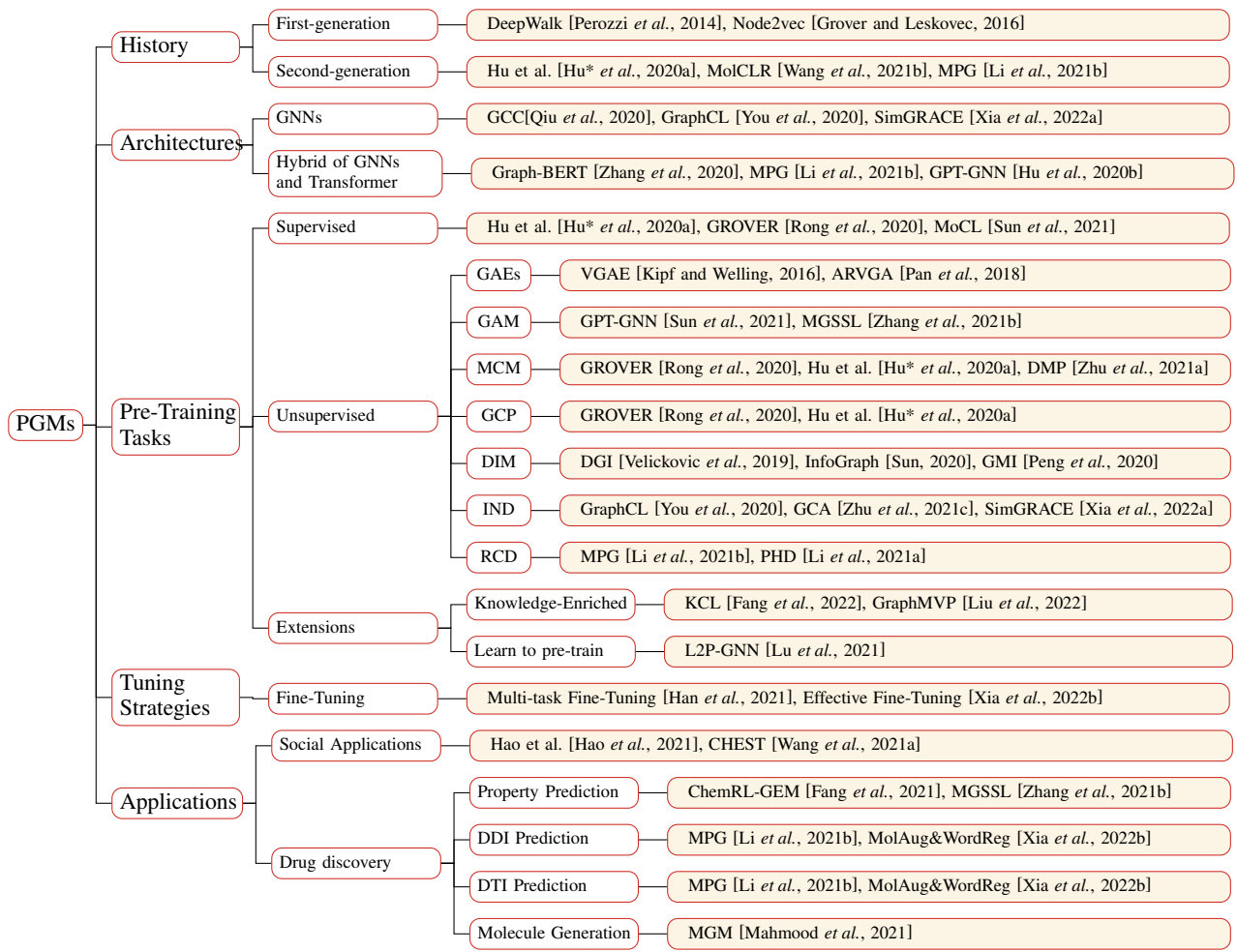
\begin{figure*}[t]
\centering
\begin{forest}
  for tree={
  forked edges,
  grow=east,
  reversed=true,
  anchor=base west,
  parent anchor=east,
  child anchor=west,
  base=middle,
  font=\footnotesize,
  rectangle,
  draw=hiddendraw,
  rounded corners,align=left,
  minimum width=2.5em,
  minimum height=1.2em,
    s sep=6pt,
    inner xsep=3pt,
    inner ysep=1pt,
  },
  where level=1{text width=4.5em}{},
  where level=2{text width=6em,font=\scriptsize}{},
  where level=3{font=\scriptsize}{},
  where level=4{font=\scriptsize}{},
  where level=5{font=\scriptsize}{},
  [PGMs
    [History,text width=4.38em
    [First-generation
        [DeepWalk~\cite{perozzi2014deepwalk}{,}
        Node2vec~\cite{grover2016node2vec},leaf,text width=29.2em]
    ]
    [Second-generation
        [Hu et al.~\cite{Hu*2020Strategies}{,}
        MolCLR~\cite{Wang2021MolCLRMC}{,}
        MPG~\cite{li2021effective}
        ,leaf,text width=29.2em]
    ]
    ]
    [Architectures,text width=4.38em
     [GNNs
        [GCC\cite{qiu2020gcc}{,}
        GraphCL~\cite{You2020GraphCL}{,}
        SimGRACE~\cite{xia2022simgrace},leaf,text width=29.2em]
     ]
    [Hybrid of GNNs \\ and Transformer, text width=6.18em
        [Graph-BERT~\cite{zhang2020graph}{,}
        MPG~\cite{li2021effective}{,}
        GPT-GNN~\cite{hu2020gpt}
        ,leaf,text width=29.2em]
    ]
    ]
    [Pre-Training\\ Tasks,text width=4.38em
      [Supervised
            [Hu et al.~\cite{Hu*2020Strategies}{,}
            GROVER~\cite{rong2020self}{,}
            MoCL~\cite{sun2021mocl}
            ,leaf,text width=29.2em]
      ]
      [Unsupervised
        [GAEs
            [VGAE~\cite{kipf2016variational}{,}  ARVGA~\cite{pan2018adversarially},leaf,text width=25.6em]
        ]
        [GAM
            [GPT-GNN~\cite{sun2021mocl}{,} MGSSL~\cite{zhang2021motif},leaf,text width=25.6em]
        ]
        [MCM
            [GROVER~\cite{rong2020self}{,} Hu et al.~\cite{Hu*2020Strategies}{,} DMP~\cite{Zhu2021DualviewMP},leaf,text width=25.6em]
        ]
        [GCP
            [GROVER~\cite{rong2020self}{,} Hu et al.~\cite{Hu*2020Strategies},leaf,text width=25.6em]
        ]
        [DIM
            [DGI~\cite{velickovic2019deep}{,} InfoGraph~\cite{Sun2020InfoGraph:}{,} GMI~\cite{peng2020graph},leaf,text width=25.6em]
        ]
        [IND
            [GraphCL~\cite{You2020GraphCL}{,}
            GCA~\cite{Zhu:2021wh}{,}
            SimGRACE~\cite{xia2022simgrace},
            ,leaf,text width=25.6em]
        ]
        [RCD
            [MPG~\cite{li2021effective}{,} PHD~\cite{ijcai2021-371},leaf,text width=25.6em]
        ]
      ]
    [Extensions
      [Knowledge-Enriched
        [
        KCL~\cite{fang2021molecular}{,}
        GraphMVP~\cite{liu2022pretraining}
        ,leaf,text width=21.6em]
      ]
      [Learn to pre-train,text width=5.98em
        [L2P-GNN~\cite{lu2021learning},leaf,text width=21.6em]
      ]
    ]
    ]
    [Tuning\\Strategies, text width=4.38em
      [Fine-Tuning
        [Multi-task Fine-Tuning~\cite{10.1145/3447548.3467450}{,} Effective Fine-Tuning~\cite{Xia2022.02.03.479055},
        leaf,text width=29.2em]
      ]
    ]
   [Applications
      [Social Applications
        [Hao et al.~\cite{hao2021pre}{,}
        CHEST~\cite{wang2021curriculum},leaf,text width=29.2em]
      ]
      [Drug discovery
        [Property Prediction,text width=5.98em
        [
        ChemRL-GEM~\cite{Fang2021geo}{,} MGSSL~\cite{zhang2021motif}
        ,leaf,text width=21.4em]
      ]
      [DDI Prediction,text width=5.98em
        [MPG~\cite{li2021effective}{,} MolAug\&WordReg~\cite{Xia2022.02.03.479055},leaf,text width=21.4em]
      ]
     [DTI Prediction,text width=5.98em
        [MPG~\cite{li2021effective}{,} MolAug\&WordReg~\cite{Xia2022.02.03.479055},leaf,text width=21.4em]
      ]
     [Molecule Generation,text width=5.98em
        [MGM~\cite{mahmood2021masked},leaf,text width=21.4em]
      ]
      ]
    ]
  ]
\end{forest}
\caption{Taxonomy of PGMs with Representative Examples.}
\label{taxonomy_of_PGMs}
\end{figure*}
\begin{itemize}
\item \emph{Comprehensive review.} To the best of our knowledge, our survey is the first work that presents a comprehensive review of PGMs. 
\item \emph{New taxonomy.}

We propose a new taxonomy, which categorizes existing PGMs from three five perspectives: (1) Brief history; (2) Model architectures; (3) Pre-training strategies; (4) Tuning Strategies; (5) Applications in social recommendation and drug discovery.
\item \emph{Abundant resources.} We collect abundant resources on PGMs, including open-sourced implementations of PGMs and paper lists\footnote{\url{https://github.com/junxia97/awesome-pre-training-on-graphs}}.
\item \emph{Future directions.} We discuss and analyze the limitations of existing PGMs. Also, we suggest possible future research directions.
\end{itemize}

\section{Notions}
Let $\mathcal{G}$ = $(\mathcal{V},\mathcal{E})$ be the graph, where $\mathcal{V} = \left\{v_{1}, v_{2}, \cdots, v_{N}\right\}, \mathcal{E} \subseteq \mathcal{V} \times \mathcal{V}$ denote the node set and edge set respectively. Besides, $\boldsymbol{X} \in \mathbb{R}^{N \times F}$ and $\boldsymbol{A} \in\{0,1\}^{N \times N}$ are the feature matrix and the adjacency matrix. $\boldsymbol{f}_{i} \in \mathbb{R}^{F}$ is the feature of $v_i$, and $\boldsymbol{A}_{i j}=1$ iff $\left(v_{i}, v_{j}\right) \in \mathcal{E}$. The objective of pre-training is to learn a generic GNN encoder $f(\boldsymbol{A}, \boldsymbol{X})$ that can deal with various downstream tasks.
\section{Overview of PGMs}
\subsection{A Brief History of PGMs for Graph}
As early as 2006, the breakthrough of deep learning came with greedy layer-wise unsupervised pre-training followed by supervised fine-tuning~\cite{hinton2006reducing}. With the development of computational power, the emergence of the deep models including GNNs and Transformer, and the constant 
enhancement of training skills, PGMs has scored remarkable progress. The development of PGMs broadly fall into two generations according to their different usage, which we will elaborate bellow. 
\subsection{First-Generation PGMs: Pre-trained Graph Embeddings}
The first-generation PGMs aim to learn good graph embeddings for various tasks such as node clustering, link prediction and visualization while these models
themselves are no longer needed by downstream tasks. Initially, inspired by Skip-Gram~\cite{mikolov2013distributed} model for word embedding, DeepWalk~\cite{perozzi2014deepwalk} pioneers graph embedding by considering the node paths traversed by random walks over graphs as the sentences and leveraging Skip-Gram for learning latent node representations. LINE~\cite{tang2015line} defines loss functions to preserve the first-order or second-order proximity separately. After optimizing the loss functions, it concatenates these representations. Following DeepWalk, Node2vec~\cite{grover2016node2vec} defines a flexible notion of a node’s network neighborhood and designs a biased random walk procedure, which efficiently explores diverse neighborhoods.  Besides, some researchers also try to learn embeddings for heterogeneous graphs, sub-graphs and molecular graph such as PTE~\cite{tang2015pte}, sub2vec~\cite{adhikari2018sub2vec}, subgraph2vec~\cite{narayanan2016subgraph2vec} and N-gram Graph~\cite{liu2019n}.
Although pre-trained graph embeddings have been shown effective in graph tasks, the learned embeddings cannot be used to initialize other models for fine-tuning over other tasks and thus impede wider applications.
\subsection{Second-Generation PGMs: Pre-trained Encoders}
With the emergence of expressive GNNs and Transformer, recent PGMs have embraced a transfer learning setting where the goal is to pre-train
a generic encoder that can deal with different tasks.
Apart from learning universal graph embeddings for downstream tasks as the first-generation PGMs, the second-generation PGMs can also provide a better model initialization, which usually leads to a better generalization performance and speeds up convergence on the target tasks.
For example, Hu et al.~\cite{Hu*2020Strategies} initialize a 5-layer Graph Isomorphism Network (GIN)~\cite{xu2018how} with the pre-trained model obtained with both graph-level and node-level pre-training tasks. Also, GCC~\cite{qiu2020gcc} utilizes a 5-layer GIN to extract representations for subgraphs and adopts subgraph discrimination in and across networks as the pre-training task to learn the intrinsic and transferable structural representations.
Since these precursor PGMs, the modern PGMs are usually trained with larger scale database, more powerful or deeper
architectures (e.g., Transformer), and new pre-training tasks. For example, the huge PGMs with ten millions of parameters have shown their powerful ability in learning universal molecular graph representations, such as  GROVER~\cite{rong2020self} and MPG~\cite{li2021effective}. Besides, various advanced pre-training tasks are proposed to capture more knowledge from database of larger scale.

\section{Model Architectures}
    The model architectures of modern PGMs broadly fall into two categories: Graph Neural Networks (GNNs), hybrid of GNNs and Transformer. We introduce them in details below. 
    \subsection{Graph Neural Networks (GNNs)}
     Recent years have witnessed the proliferation of GNNs. The structure of graph data guides the aggregation of local neighborhood information and lead to a more contextual representation for each node. Also, we can adopt a graph pooling operation~\cite{rethinkpooling2020} to get the representation for the whole graph.
    %  Formally, suppose $H_{v}^{(l)}$ is the node representation of node $v$ at the $l$-th GNN layer and $N(v)$ is all the source nodes of node $v$, the update procedure from the $(l-1)$-th layer to the $(l)$-th layer is:
    % \begin{equation}
    % H_{v}^{(l)} \leftarrow \underset{\forall u\in N(v), \forall e \in E(u, v)}{\textbf{Aggregate }}\left(\left\{\operatorname{\textbf{Combine}}\left(H_{u}^{(l-1)} ; H_{v}^{(l-1)}, e\right)\right\}\right)
    % \end{equation}
    % where $E(u, v)$ denotes all the edges from node $u$ to $v$. \textbf{Combine($\cdot$)} combines the information of neighbor and \textbf{Aggregate($\cdot$)} serves as the aggregation function of the neighborhood information. After $L$ iterations of message passing, the hidden states $H_{v}^{(L)}$  in the last iteration are used as the embeddings of $v$. Finally, we can adopt a \textbf{Readout($\cdot$)} operation to get the representation for the whole graph $G$:
    % \begin{equation}
    % \mathbf{h}_{G}=\textbf{Readout}\left(\left\{\mathbf{h}_{v}^{\left(0\right)}, \ldots, \mathbf{h}_{v}^{\left(L\right)} \mid v \in \mathcal{V}\right\}\right),
    % \end{equation}
    % where $\mathcal{V}$ is the node set of graph $G$. 
    For PGMs, GIN~\cite{xu2018how} is the most popular encoder for its high expressive power~\cite{Hu*2020Strategies}. Besides, Heterogeneous Attention Network (HAN)~\cite{wang2019heterogeneous} is a more suitable alternative for pre-training on heterogeneous graphs~\cite{heco,Zhu:2020ui}. 
    % HGT~\cite{hu2020heterogeneous} and HAN~\cite{wang2019heterogeneous} are more suitable for for heterogeneous graphs. In addition, common GNNs including GCN~\cite{kipf2017semi-supervised}, GAT~\cite{velickovic2018graph}, GraphSAGE~\cite{hamilton2017inductive} are adopted in some ablation studies on the base GNN for some proposed pre-training methods.
    \subsection{Hybrid of GNNs and Transformer}
    To leverage the high expressiveness of Transformer, several recent works try to integrate GNNs into Transformer-style models. 
%     To start, we briefly introduce Transformer architecture which is composed of a stack of identical Transformer layers. Each layer consists of a multi-head attention module (MHA) followed by a feed-forward module (FFN), with a residual connection around each. The vanilla single-head attention operates as
% $$
% \operatorname{Att}(Q, K, V)=\operatorname{Softmax}\left(\frac{Q K^{\top}}{\sqrt{d}}\right) V,
% $$
% where $Q, K, V \in \mathbb{R}^{l \times d}$ are $d$-dimensional vector representations of $l$ words in sequences of queries, keys and values. In MHA, the $h$-th attention head is parameterized by $W_{h}^{Q}, W_{h}^{K}, W_{h}^{V} \in \mathbb{R}^{d \times d_{h}}$ as
% $$
% \mathrm{H}_{h}\left(\boldsymbol{q}, \boldsymbol{x}, W_{h}^{\{Q, K, V\}}\right)=\operatorname{Att}\left(\boldsymbol{q} W_{h}^{Q}, \boldsymbol{x} W_{h}^{K}, \boldsymbol{x} W_{h}^{V}\right),
% $$
% where $\boldsymbol{q} \in \mathbb{R}^{l \times d}$ and $\boldsymbol{x} \in \mathbb{R}^{l \times d}$ are the query and key/value vectors. In M``HA, $H$ independently parameterized attention heads are applied in parallel, and the outputs are aggregated by $W_{h}^{O} \in \mathbb{R}^{d_{h} \times d}$ :
% $$
% \operatorname{MHA}(\boldsymbol{q}, \boldsymbol{x})=\sum_{h}^{H} \mathrm{H}_{h}\left(\boldsymbol{q}, \boldsymbol{x}, W_{h}^{\{Q, K, V\}}\right) W_{h}^{O}
% $$
% Each FFN module contains a two-layer fully connected network. Given the input embedding $z$, we let $\operatorname{FFN}(\boldsymbol{z})$ denote the output of a FFN module. 
For PGMs, GROVER~\cite{rong2020self} first utilize GNNs to capture local
structural information of the graph data and then the outputs of the GNNs as queries, keys and values for Transformer encoder. They claim that this bi-level information extraction strategy largely enhances the representational power of the proposed models. Analogously, MPG~\cite{li2021effective} devises a neighbor attention module to get produce a message representation for each node and feed it to a fully
connected feed-forward network. With the proper message representation obtained, they adopt a GRU network~\cite{bff0e6bd8f4a4f0d9735bf1728fb43ef} to update node representation. For heterogeneous graphs pre-training~\cite{hu2020gpt,jiang2021contrastive}, Heterogeneous Graph Transformer (HGT)~\cite{hu2020heterogeneous} is usually adopted as the encoder. 

\section{Pre-training Strategies}
In this section, we will elaborate on the supervised, unsupervised pre-training strategies and the extensions of PGMs. 
\renewcommand\arraystretch{1.38}
\begin{table*}[t]
\caption{Loss functions of various unsupervised pre-training strategies}
\label{Table_loss}
\centering
\footnotesize
\begin{tabular}{llp{0.518\textwidth}}
\toprule \textbf{Task} & \textbf{Loss Function} & \textbf{Description} \\
\hline GAEs & $\mathcal{L}_{\mathrm{GAEs}}=-\log p\left(\boldsymbol{X}, \mathcal{E} \mid \mathcal{G}\right)$ & Graph construction.  \\
 GAM & $\mathcal{L}_{\mathrm{GAM}}=-\sum_{i=1}^{|\mathcal{V}|} \log p\left(\boldsymbol{X}_{i}, \mathcal{E}_{i} \mid \boldsymbol{X}_{<i}, \mathcal{E}_{<i}\right)$ & $\boldsymbol{X}_{<i}, \mathcal{E}_{<i}$ are the attributes and edges generated before node $i$ respectively.\\
 MCM & $\mathcal{L}_{\mathrm{MCM}}=-\sum_{\widehat{\mathcal{G}} \in m(\mathcal{G})} \log p\left(\widehat{\mathcal{G}} \mid \mathcal{G}_{\backslash m(\mathcal{G})}\right)$ & $m(\mathcal{G})$ are the masked components from $\mathcal{G}$ and $\mathcal{G}_{\backslash m(\mathcal{G})}$ are the rest.
 \\
 GCP & $\mathcal{L}_{\mathrm{GCP}}=-\log p(t \mid \mathcal{G}_1, \mathcal{G}_2)$ & $t=1$ if neighborhood graph $\mathcal{G}_1$ and contexts $\mathcal{G}_2$ belong to the same node. \\
 IND & $\mathcal{L}_{\mathrm{IND}}=-s\left(\mathcal{G}, \mathcal{G}^{+}\right)+\log \sum_{\mathcal{G}^{-} \in \mathcal{N}} s\left(\mathcal{G},\mathcal{G}^{-}\right)$ &$\mathcal{N}$ is a set of negatives; $\mathcal{G}^{+}$ is a positive sample.\\
DIM & $\mathcal{L}_{\mathrm{IND}}=-s\left(\mathcal{G}, \mathcal{C}\right)+\log \sum_{\mathcal{C}^{-} \in \mathcal{N}} s\left(\mathcal{G},\mathcal{C}^{-}\right)$ &$\mathcal{N}$ is a set of negatives; $\mathcal{C}$ is a substructure of $\mathcal{G}$.\\
 RCD & $\mathcal{L}_{\mathrm{RCD}}=-\log p(t \mid \mathcal{G}_1, \mathcal{G}_2)$ & $t=1$ if two half graphs $\mathcal{G}_1$ and $\mathcal{G}_2$ are homologous couples.\\
\toprule
\end{tabular}
\end{table*}

    \subsection{Supervised Strategies}
    Although the supervised labels are often time-consuming and expensive to collect, some cheaper annotations that may be less related to downstream tasks can also help pre-training on graphs, especially in biochemical domains. For example, Hu et al. ~\cite{Hu*2020Strategies} propose to pre-train GNNs to predict essentially all the properties of molecules that have been experimentally measured so far. Analogously, for protein function prediction, they pre-train GNNs to predict the existence of diverse protein functions that have been validated so far. Also, they leave a future work to take the structural similarities between two graphs as supervision. Inspired by this, MoCL~\cite{sun2021mocl} first calculates the Tanimoto coefficient~\cite{Bajusz2015WhyIT} between of two molecules as the measure of structural similarity, which serves as the supervisions for the pre-training. For molecular graphs, one important class of motifs in molecules are functional groups that encodes rich domain knowledge of molecules and can be easily detected by the professional software such as RDkit\footnote{\url{https://www.rdkit.org/}} or developed algorithms~\cite{ertl2017algorithm}. In light of this, GROVER~\cite{rong2020self} and MGSSL~\cite{zhang2021motif} propose to predict the presence of the motifs or generate the motifs respectively. Although the supervised pre-trainings brings remarkable improvements, they often require domain-specific knowledge which significantly limits their wider application. More importantly, some supervised pre-training tasks might be unrelated to the downstream task of interest and can even hurt the downstream performance.

\subsection{Unsupervised Strategies}
    \subsubsection{Graph AutoEncoders (GAEs)}
    Graph reconstruction serves as a natural self-supervision for learning discriminative representations. The prediction targets in graph reconstruction are certain parts of the given graphs such as the attribute of a subset of nodes or the existence of edge between a pair of nodes. Inspired by the success of AutoEncoders in CV and NLP, various GAEs have been proposed recently. Among many, GAE~\cite{kipf2016variational} is the simplest version of the graph autoencoders, which reconstructs adjacency matrix $\widehat{\boldsymbol{A}}$ with,
    \begin{equation}
            \begin{aligned} \widehat{\boldsymbol{A}} &=\sigma\left(\boldsymbol{H} \boldsymbol{H}^{T}\right), \boldsymbol{H} &=f(\boldsymbol{A}, \boldsymbol{X}), \end{aligned}
    \end{equation}
    and is optimized by the binary cross-entropy loss between $\widehat{\boldsymbol{A}}$ and $\boldsymbol{A}$. $\sigma(\cdot)$ is the sigmoid function. Also, there exist multiple variants of GAEs that utilize graph reconstruction to pre-train the GNNs. Representative examples include VGAE~\cite{kipf2016variational}, MGAE~\cite{wang2017mgae}, ARVGA~\cite{pan2018adversarially}, SIG-VAE~\cite{hasanzadeh2019semi} and so on.
    % \subsubsection{Graph Context Prediction (GCP)}
    \subsubsection{Graph Autoregressive Modeling (GAM)}
    Following the idea of GPT that conducts generative language model pre-training, GPT-GNN~\cite{hu2020gpt} proposes an autoregressive framework to perform reconstruction on given graphs iteratively, which is different from graph autoencoders that reconstruct the graph all at once. In particular, given a graph with
    its nodes and edges randomly masked, GPT-GNN generates one masked node and its edges at a time and optimizes the parameterized models via maximizing the likelihood of the node and edges generated in the current iteration. Then, it iteratively generates nodes and edges
    until all masked nodes are generated. Analogously, MGSSL~\cite{zhang2021motif} generates molecular graph motifs in an autoregressive way based on existing motifs and connections. 
    \subsubsection{Masked Components Modeling (MCM)}
    Similar to masked language modeling (MLM) that masks out some tokens from the
 input sentences and then trains the model to predict the masked tokens by the rest of the tokens, MCM first mask out some component from the graphs and then trains the model to predict them. For example, Hu et.al~\cite{Hu*2020Strategies} propose attribute masking where the input node/edge attributes are randomly masked, and the GNN is asked to predict them. Also, GROVER~\cite{rong2020self} tries to predict the masked subgraphs to capture the contextual information in molecular graphs.  These masking methods are especially beneficial for richly-annotated graphs from scientific domains. For example, masking nodes attributes (atom type) enables GNNs to learn simple
chemistry rules such as valency, as well as potentially more complex chemistry phenomenon such as the electronic or steric properties of functional groups.
    \subsubsection{Graph Context Prediction (GCP)}
   GCP is proposed to explore the distribution of graph structure in graph data. For example, Hu et al.~\cite{Hu*2020Strategies} use subgraphs to predict their surrounding graph structures. They pre-train a GNN so that it maps nodes appearing in similar structural contexts to nearby embeddings. GROVER tries to predict the context-aware properties of the target node/edge within some local subgraph. Here, the properties refer to some node-edge counts terms around the target node/edge. 
    \subsubsection{Graph Contrastive Learning (GCL)}
    \paragraph{Deep InfoMax (DIM)}
    Deep InfoMax is originally proposed for images, which improves the quality of the
    representation by maximizing the mutual information between an image representation and local regions of the image. For graphs, initially, DGI~\cite{velickovic2019deep} and InfoGraph~\cite{Sun2020InfoGraph:} are proposed to obtain expressive representations for graphs or nodes via maximizing the mutual information between graph-level representations and substructure-level representations of different granularity. Similarly, GMI~\cite{peng2020graph} adopts two discriminators to directly measure mutual information between input and representations of both nodes and edges. Besides, MVGRL~\cite{hassani2020contrastive} performs node diffusion to generate augmented view and then maximizes the mutual information between original and augmented views by contrasting node representations of one view with graph representation of the other view and vice versa.

    \paragraph{Instance Discrimination (IND)}
    IND is one of the most popular pre-training tasks which embeds augmented versions of the anchor close to each other (positive samples) and pushes the embeddings of other samples (negatives) apart. For node-level representations, GRACE~\cite{Zhu:2020vf} and its variants~\cite{Zhu:2021wh,Jin2021MultiScaleCS,xia2021debiased} maximize the agreement of node embeddings across two corrupted views of the graph. Besides, GraphCL~\cite{You2020GraphCL} and its variants~\cite{you2021graph,sun2021mocl,suresh2021adversarial} propose various advanced augmentations strategies for graph-level pre-training. More recently, some works such as BGRL~\cite{thakoor2021bootstrapped}, CCA-SSG~\cite{zhang2021canonical}, LP-Info~\cite{you2022bringing} and SimGRACE~\cite{xia2022simgrace} try to simplify graph contrastive learning via discarding the negatives, parameterized mutual information estimator or even data augmentations respectively. We develop an open-source graph contrastive learning (GCL) library\footnote{https://github.com/GraphCL/PyGCL} for PyTorch~\cite{zhu2021an}.
    \subsubsection{Replaced Component Detection (RCD)}
    To capture the global information of graphs, RCD is proposed as a graph-level pre-training task
    on a random permutation of input graphs. For example, PHD~\cite{ijcai2021-371} first decomposes each molecular graph in the database into two half-graphs and replace one of them with a half-graph from other graph randomly. The GNN encoder is pre-trained to detect whether two half-graphs are homologous couples.
\renewcommand\arraystretch{1.0}
\begin{table*}[t]
\caption{List of Representative PGMs. Here KG of KCL is  Chemical Element Knowledge Graph.}
\label{Table_PGMs}
\centering
\scriptsize
\begin{tabular}{l|lllp{0.188\textwidth}p{0.060\textwidth}}
\toprule \textbf{PGMs} & \textbf{Input} & \textbf{Architecture} &\textbf{Pre-Training Task} & \textbf{Pre-training Database}   & \textbf{\# Params.}  \\
\hline Hu et al.~\cite{Hu*2020Strategies}  &Graph&5-layer GIN  & GCP + MCM & ZINC15 (2M) + ChEMBL (456K) &$\sim$ 2M \\
Graph-BERT~\cite{zhang2020graph} &Graph & Graph Transformer~\cite{zhang2020graph}  & GAEs & Cora + CiteSeer + PubMed & N/A \\
GPT-GNN~\cite{hu2020gpt}  &Graph& HGT~\cite{hu2020heterogeneous}  & GAM & OAG + Amazon  & N/A  \\
GCC~\cite{qiu2020gcc} &Graph  &5-layer GIN  & IND & Academia + DBLP + IMDB + Facebook + LiveJournal  & \textless1M\\
GraphCL~\cite{You2020GraphCL}  &Graph &5-layer GIN &IND  & ZINC15 (2M) + ChEMBL (456K) & $\sim$ 2M \\
JOAO~\cite{you2021graph} &Graph  &5-layer GIN  &IND  & ZINC15 (2M) + ChEMBL (456K) & $\sim$ 2M\\
AD-GCL~\cite{suresh2021adversarial} &Graph  & 5-layer GIN &IND  & ZINC15 (2M) + ChEMBL (456K) & $\sim$ 2M \\
GraphLog~\cite{xu2021self} &Graph  & 5-layer GIN &IND  & ZINC15 (2M) + ChEMBL (456K) &$\sim$ 2M \\
GROVER~\cite{rong2020self} &Graph & GTransformer~\cite{rong2020self}& GCP + MCM  & ZINC + ChEMBL (10M)  & 48M$\sim$100M  \\
MGSSL~\cite{zhang2021motif}&Graph &5-layer GIN  &MCM + GAM  & ZINC15 (250K) &$\sim$ 2M \\
CPT-HG~\cite{jiang2021contrastive} &Graph  &HGT~\cite{hu2020heterogeneous} &IND  &DBLP + YELP + Aminer   & N/A  \\
PGM~\cite{li2021effective} &Graph  &MolGNet~\cite{li2021effective} & RCD + MCM & ZINC + ChEMBL (11M)  & 53M \\
LP-Info~\cite{you2022bringing} &Graph  &5-layer GIN &IND  & ZINC15 (2M) + ChEMBL (456K) & $\sim$ 2M \\
SimGRACE~\cite{xia2022simgrace} &Graph &5-layer GIN&IND  & ZINC15 (2M) + ChEMBL (456K) & $\sim$ 2M\\
\midrule
MolCLR~\cite{Wang2021MolCLRMC} &Graph + SMILES & GCN + GIN & IND & PubChem (10M) &N/A\\
DMP~\cite{Zhu2021DualviewMP} &Graph + SMILES & DeeperGCN + Transformer & MCM + IND & PubChem (110M) &104.1 M\\
ChemRL-GEM~\cite{Fang2021geo} & Graph + Geometry & GeoGNN~\cite{Fang2021geo} &MCM+GCP  & ZINC15 (20M)  & N/A \\
KCL~\cite{fang2021molecular} & Graph + KG  & GCN + KMPNN~\cite{fang2021molecular} &IND  & ZINC15 (250K)  & \textless 1M  \\
3D Infomax~\cite{stark20213d} & 2D and 3D graph   &  PNA~\cite{corso2020principal}& IND & QM9(50K) + GEOM-drugs(140K) + QMugs(620K)  & N/A  \\
GraphMVP~\cite{liu2022pretraining} & 2D and 3D graph &5-layer GIN + SchNet~\cite{NIPS2017_303ed4c6} & IND + GAEs & GEOM (50k)  & $\sim$ 2M  \\
\toprule
\end{tabular}
\end{table*}
    \subsection{Extensions}
        \subsubsection{Knowledge-Enriched Pre-training}
        PGMs usually learn universal graph representation from general-purpose graphs database. However, they often lack domain-specific knowledge. To enhance their performance, several recent works try to inject external knowledge during pre-taining. For example, GraphCL~\cite{You2020GraphCL} first pointed out that edge perturbation is conceptually incompatible with domain knowledge and empirically unhelpful for down-stream performance for chemical compounds. And then, they avoid adopting edge perturbation for molecular graphs augmentation. To incorporate the domain knowledge into pre-training more explicitly, MoCL~\cite{sun2021mocl} proposed a new augmentation operator called substructure substitution, in which a valid substructure in a molecule is replaced by a bioisostere~\cite{meanwell2011synopsis} which produces a new molecule with similar physical or chemical properties as the original one. They compile 230 substitution rules from domain resource in total and emperically validate their effectiveness. More recently, to capture the correlations between atoms that have common attributes but are not directly connected by bonds, KCL~\cite{fang2021molecular} construct a Chemical Element Knowledge Graph (KG) to summarize microscopic associations between elements and propose a novel Knowledge-enhanced Contrastive Learning (KCL) framework for molecular representation learning. Considering that 3D geometric information of molecule also plays a vital role in predicting molecular functionalities, 3DInfoMax~\cite{stark20213d} proposes pre-training a model to reason about the geometry of molecules given only their 2D molecular graphs while GraphMVP~\cite{liu2022pretraining} performs self-supervised pre-training via maximizing the correspondence and consistency between 2D topological structures and 3D geometric views. 
        \subsubsection{Learn to Pre-train}
        Due to the the divergence of the optimization objectives between pre-training and fine-tuning steps, there exists a gap between them which will significantly hurt the
        generalization ability of PGMs. To narrow this gap, L2P-GNN~\cite{lu2021learning} simulate the fine-tuning via creating new tasks during pre-training. This setup enables PGMs to adapt to new tasks quickly and lead to better generalization on downstream tasks.

\section{Tunnig Strategies}
        Although PGMs can capture abuntant knowledge that is useful for downstream tasks, the process of vanilla fine-tuning is still brittle. For example, Xia et al.~\cite{Xia2022.02.03.479055} observe that PGMs are prone to overfit insufficient labeled data for downstream tasks due to PGMs' high complexity. In particular, unlike image or text data, getting labels for biochemical graph data often requires wet-lab experiments which often lead to inaccurate annotations~\cite{xia2021towards}. To enrich the labeled data of downstream tasks, they propose to augment molecular graph data with chemical enantiomers and homologies, which share the similar physical (permeability, solubilityand etc) or chemical (toxicity, side effect and ect) properties with original molecules. To control the complexity of PGMs, they introduce a new regularization built on dropout which encourages the output of PGMs not to change much when injecting a small perturbation and thus effectively controls PGMs' capacity. Besides, catastrophic forgetting often happens when adapting PGMs to downstream tasks. In other words, PGMs often forget their learned general knowledge when fine-tuning. To alleviate this issue, Han et al.~\cite{10.1145/3447548.3467450} utilize meta learning to adaptively select and combine various auxiliary tasks with the target task in fine-tuning stage to achieve a better adaptation. This preserves sufficient knowledge captured by self-supervised auxiliary tasks while improving the effectiveness of transfer learning on GNNs. However, it takes the auxiliary tasks of pre-training as a prerequisite, which impedes the usage of their methods in practice where the auxiliary tasks are often unknown.  

\section{Applications}
    \subsection{Social Recommendations}
    Owing to the outstanding performance in graph data learning, GNNs have been widely applied to recommender systems. Despite their proliferation, GNNs based recommender systems are still fraught with issues. For example, cold-start problem impedes their wider applications because GNNs fail to learn high-quality embeddings for the cold-start users/items with sparse interactions. To alleviate this critical issue, Hao et al.~\cite{hao2021pre} propose to pre-train the GNN model via predicting the ground-truth embeddings of users/item. In this way, they can enhance the embeddings of the cold-start users or items with the PGMs. For Heterogeneous Information Network (HIN)-based recommendation~\cite{hu2018leveraging,jin2020efficient}, Wang et al.~\cite{wang2021curriculum} capture the rich semantics in the subgraph extracted from HIN via a heterogeneous subgraph Transformer and devise a curriculum pre-training strategy to provide an elementary-to-advanced learning process. 
This strategy can smoothly transfer basic semantics in HIN for modeling user-item interaction relation and effectively utilize the rich information in HIN for recommendation task.
    \subsection{Drug Discovery}
     A molecule is naturally treated as a graph, where nodes refer to atoms and edges correspond to chemical bonds. Recently, the advancements in graph pre-training provide opportunities to expedite drug discovery and development pipeline.  
    For example, the oral bioavailability of a drug is related to many properties, such as solubility in gastrointestinal tract, intestinal membrane permeability and intestinal/hepatic first-pass metabolism~\cite{hou2007adme}. However, it is often time-consuming and unsafe to conduct such experiments on human bodies. In \emph{molecular property prediction} tasks, PGMs can be directly applied as a drug encoder to obtain expressive representations~\cite{Wang2021MolCLRMC,rong2020self}. Besides, \emph{drug-drug interaction (DDI) prediction} is also of vital importance in drug discovery because DDIs may lead to adverse drug reactions which will damage the health or even cause death. DDI prediction tasks can be regarded as a task that classify the influence of combining drugs into three categories: synergistic, additive and antagonistic. Works on molecular graph pre-training, such as MPG~\cite{li2021effective} and WordReg \& MolAug~\cite{Xia2022.02.03.479055}, have applied DDI prediction as a downstream task to reveal the effectiveness of the PGMs. Also, \emph{drug-target interaction (DTI) prediction} is important in drug discovery. When a new indication occurs, the best choice for coping is to recycle approved drugs because
    of their availability and known safety profiles. In this case, PGMs can be directly applied as a drug encoder. Thus, the well pre-trained model weights can be regarded as the initial weights of drug encoder. The drug encoder and target encoder are then trained with the DTI prediction task. Related works include MPG and WordReg \& MolAug have followed this idea to achieve DTI prediction. For \emph{molecule generation}, MGM~\cite{mahmood2021masked} introduces a masked graph model, which learns a distribution over graphs by capturing conditional distributions over unobserved nodes (atoms) and edges (bonds) given observed ones. 
    %  \subsubsection{Molecular property prediction}
    %  The oral bioavailability of a drug is related to many properties, such as solubility in gastrointestinal tract, intestinal membrane permeability and intestinal/hepatic first-pass metabolism~\cite{hou2007adme}. However, it is often time-consuming and unsafe to conduct such experiments on human bodies. Therefore, .
    %  \subsubsection{Drug–drug interaction prediction (DDI)}
    %  \subsubsection{Drug–target interaction prediction (DTI)}
\section{Conclusion and Future Outlooks}
Despite the fruitful progress of PGMs, challenges still exist due to the complexity of graph data. In this section, we suggest several promising research directions for future.
\subsection{Better Knowledge Transfer}
Currently, tremendous efforts are focusing on pre-training strategies. However, how to leverage these huge PGMs is still under-explored compared to PLMs. Fine-tuning is an dominant technique to adapt the knowledge to various downstream tasks, but there are several nonnegligible deficiencies to be solved. The first one is poor generalization of PGMs especially for various molecular tasks where collecting labeled data is laborious. The second issue is parameter inefficiency. The fine-tuned parameters vary across both datasets and tasks, which are often huge in scale and thus being inconvenient in special scenarios such as low-capacity devices. Furthermore, there are some promising alternatives to mine the knowledge from PGMs. For example, PGMs can serve as the feature extractor to extract expressive representations as adopted in graph self-supervised learning. Also, distilling the knowledge from PGMs as adopted in NLP is expected~\cite{textbrewer-acl2020-demo}.

\subsection{Interpretability of PGMs}
Despite their proliferation, a major limitation of PGMs is that they are not amenable to interpretability. Worse still, unlike CNNs for images, interpreting PGMs is more difficult due to the complexities of both the Transformer-style architecture and graph data. However, for some specific scenarios like molecular toxicity prediction, it is of vital importance for the PGMs possess the ability to explain the reason why a molecule is non-toxic. Also, interpretability can accelerate some scientific findings such as identifying biomarks. Overall, as a key component in graph-related applications, the interpretability of PGMs remain to be explored further in many respects, which helps us understand how PGMs work and provides a guide for better usage and further improvement.
\subsection{Broader Scope of Applications}
PGMs have been applied in various sub-tasks in drug discovery such as molecular property prediction, DDI and DTI. However, it is still underexplored how PGMs can benefit other small molecule-related tasks such as chemical reaction prediction, retrosynthesis, de novo molecule design and optimization. For macromolecules such as proteins, recent works demonstrate that GNNs can help learn expressive representations for them~\cite{xia2021geometric}. More endeavors are still expected to study whether PGMs are conducive to protein representation learning.  
\setlength{\baselineskip}{0.5pt}
% \small
\bibliographystyle{named}
\bibliography{ijcai22}

\begin{thebibliography}{}

\bibitem[\protect\citeauthoryear{B.~Adhikari and
  Prakash}{2018}]{adhikari2018sub2vec}
N.~Ramakrishnan B.~Adhikari, Y.~Zhang and B.~Prakash.
\newblock Sub2vec: Feature learning for subgraphs.
\newblock In {\em PAKDD}, 2018.

\bibitem[\protect\citeauthoryear{Bajusz \bgroup \em et al.\egroup
  }{2015}]{Bajusz2015WhyIT}
D{\'a}vid Bajusz, Anita R{\'a}cz, and K{\'a}roly H{\'e}berger.
\newblock Why is tanimoto index an appropriate choice for fingerprint-based
  similarity calculations?
\newblock {\em Journal of Cheminformatics}, 7, 2015.

\bibitem[\protect\citeauthoryear{Cho \bgroup \em et al.\egroup
  }{2014}]{bff0e6bd8f4a4f0d9735bf1728fb43ef}
Kyunghyun Cho, B~{van Merrienboer}, et~al.
\newblock Learning phrase representations using rnn encoder-decoder for
  statistical machine translation.
\newblock In {\em EMNLP}, 2014.

\bibitem[\protect\citeauthoryear{Corso \bgroup \em et al.\egroup
  }{2020}]{corso2020principal}
G.~Corso, L.~Cavalleri, et~al.
\newblock Principal neighbourhood aggregation for graph nets.
\newblock {\em NeurIPS}, 2020.

\bibitem[\protect\citeauthoryear{Devlin \bgroup \em et al.\egroup
  }{2019}]{devlin2019bert}
Jacob Devlin, Ming-Wei Chang, and others.
\newblock Bert: Pre-training of deep bidirectional transformers for language
  understanding.
\newblock {\em NACCL}, 2019.

\bibitem[\protect\citeauthoryear{Ertl}{2017}]{ertl2017algorithm}
P.~Ertl.
\newblock An algorithm to identify functional groups in organic molecules.
\newblock {\em Journal of cheminformatics}, 2017.

\bibitem[\protect\citeauthoryear{Fang \bgroup \em et al.\egroup
  }{2021}]{Fang2021geo}
X.~Fang, L.~Liu, and others.
\newblock Geometry-enhanced molecular representation learning for property
  prediction.
\newblock {\em Nature Machine Intelligence}, 2021.

\bibitem[\protect\citeauthoryear{Fang \bgroup \em et al.\egroup
  }{2022}]{fang2021molecular}
Yin Fang, Qiang Zhang, and others.
\newblock Molecular contrastive learning with chemical element knowledge graph.
\newblock {\em AAAI}, 2022.

\bibitem[\protect\citeauthoryear{Grover and
  Leskovec}{2016}]{grover2016node2vec}
A.~Grover and J.~Leskovec.
\newblock node2vec: Scalable feature learning for networks.
\newblock In {\em KDD}, 2016.

\bibitem[\protect\citeauthoryear{Han \bgroup \em et al.\egroup
  }{2021}]{10.1145/3447548.3467450}
X.~Han, Z.~Huang, and others.
\newblock Adaptive transfer learning on graph neural networks.
\newblock In {\em KDD}, 2021.

\bibitem[\protect\citeauthoryear{Hao \bgroup \em et al.\egroup
  }{2019}]{hao2019visualizing}
Yaru Hao, Li~Dong, and others.
\newblock Visualizing and understanding the effectiveness of bert.
\newblock {\em EMNLP/IJCNLP}, 2019.

\bibitem[\protect\citeauthoryear{Hao \bgroup \em et al.\egroup
  }{2021}]{hao2021pre}
B.~Hao, J.~Zhang, and others.
\newblock Pre-training graph neural networks for cold-start users and items
  representation.
\newblock In {\em WSDM}, 2021.

\bibitem[\protect\citeauthoryear{Hasanzadeh \bgroup \em et al.\egroup
  }{2019}]{hasanzadeh2019semi}
A.~Hasanzadeh, E.~Hajiramezanali, and others.
\newblock Semi-implicit graph variational auto-encoders.
\newblock {\em NeurIPS}, 2019.

\bibitem[\protect\citeauthoryear{Hassani and
  Khasahmadi}{2020}]{hassani2020contrastive}
K.~Hassani and A.~Khasahmadi.
\newblock Contrastive multi-view representation learning on graphs.
\newblock In {\em ICML}, 2020.

\bibitem[\protect\citeauthoryear{Hinton and
  Salakhutdinov}{2006}]{hinton2006reducing}
Geoffrey~E Hinton and Ruslan~R Salakhutdinov.
\newblock Reducing the dimensionality of data with neural networks.
\newblock {\em science}, 2006.

\bibitem[\protect\citeauthoryear{Hou \bgroup \em et al.\egroup
  }{2007}]{hou2007adme}
T.~Hou, J.~Wang, W.~Zhang, and X.~Xu.
\newblock Adme evaluation in drug discovery. 6. can oral bioavailability in
  humans be effectively predicted by simple molecular property-based rules?
\newblock {\em Journal of Chemical Information and Modeling}, 2007.

\bibitem[\protect\citeauthoryear{Hu \bgroup \em et al.\egroup
  }{2018}]{hu2018leveraging}
B.~Hu, C.~Shi, W.~Zhao, and P.~Yu.
\newblock Leveraging meta-path based context for top-n recommendation with a
  neural co-attention model.
\newblock In {\em KDD}, 2018.

\bibitem[\protect\citeauthoryear{Hu* \bgroup \em et al.\egroup
  }{2020a}]{Hu*2020Strategies}
Weihua Hu*, Bowen Liu*, and others.
\newblock Strategies for pre-training graph neural networks.
\newblock In {\em ICLR}, 2020.

\bibitem[\protect\citeauthoryear{Hu \bgroup \em et al.\egroup
  }{2020b}]{hu2020gpt}
Z.~Hu, Y.~Dong, and others.
\newblock Gpt-gnn: Generative pre-training of graph neural networks.
\newblock In {\em KDD}, 2020.

\bibitem[\protect\citeauthoryear{Hu \bgroup \em et al.\egroup
  }{2020c}]{hu2020heterogeneous}
Z.~Hu, Y.~Dong, K.~Wang, and Y.~Sun.
\newblock Heterogeneous graph transformer.
\newblock In {\em WWW}, 2020.

\bibitem[\protect\citeauthoryear{Jiang \bgroup \em et al.\egroup
  }{2021}]{jiang2021contrastive}
Xunqiang Jiang, Yuanfu Lu, and others.
\newblock Contrastive pre-training of gnns on heterogeneous graphs.
\newblock In {\em CIKM}, 2021.

\bibitem[\protect\citeauthoryear{Jin \bgroup \em et al.\egroup
  }{2020}]{jin2020efficient}
Jiarui Jin, Jiarui Qin, and others.
\newblock An efficient neighborhood-based interaction model for recommendation
  on heterogeneous graph.
\newblock In {\em KDD}, 2020.

\bibitem[\protect\citeauthoryear{Jin \bgroup \em et al.\egroup
  }{2021}]{Jin2021MultiScaleCS}
M.~Jin, Y.~Zheng, and others.
\newblock Multi-scale contrastive siamese networks for self-supervised graph
  representation learning.
\newblock In {\em IJCAI}, 2021.

\bibitem[\protect\citeauthoryear{Kipf and Welling}{2016}]{kipf2016variational}
Thomas~N Kipf and Max Welling.
\newblock Variational graph auto-encoders.
\newblock {\em arXiv:1611.07308}, 2016.

\bibitem[\protect\citeauthoryear{Li \bgroup \em et al.\egroup
  }{2021a}]{ijcai2021-371}
P.~Li, J.~Wang, and others.
\newblock Pairwise half-graph discrimination: A simple graph-level
  self-supervised strategy for pre-training graph neural networks.
\newblock In {\em IJCAI}, 2021.

\bibitem[\protect\citeauthoryear{Li \bgroup \em et al.\egroup
  }{2021b}]{li2021effective}
Pengyong Li, Jun Wang, and others.
\newblock An effective self-supervised framework for learning expressive
  molecular global representations to drug discovery.
\newblock {\em BIB}, 2021.

\bibitem[\protect\citeauthoryear{Liu \bgroup \em et al.\egroup
  }{2019}]{liu2019n}
Shengchao Liu, Mehmet~F Demirel, and Yingyu Liang.
\newblock N-gram graph: Simple unsupervised representation for graphs, with
  applications to molecules.
\newblock {\em NeurIPS}, 2019.

\bibitem[\protect\citeauthoryear{Liu \bgroup \em et al.\egroup
  }{2021}]{liu2021graph}
Yixin Liu, Shirui Pan, and others.
\newblock Graph self-supervised learning: A survey.
\newblock {\em arXiv:2103.00111}, 2021.

\bibitem[\protect\citeauthoryear{Liu \bgroup \em et al.\egroup
  }{2022}]{liu2022pretraining}
Shengchao Liu, Hanchen Wang, and others.
\newblock Pre-training molecular graph representation with 3d geometry.
\newblock In {\em ICLR}, 2022.

\bibitem[\protect\citeauthoryear{Lu \bgroup \em et al.\egroup
  }{2021}]{lu2021learning}
Y.~Lu, X.~Jiang, Y.~Fang, and C.~Shi.
\newblock Learning to pre-train graph neural networks.
\newblock In {\em AAAI}, 2021.

\bibitem[\protect\citeauthoryear{Mahmood \bgroup \em et al.\egroup
  }{2021}]{mahmood2021masked}
O.~Mahmood, E.~Mansimov, and others.
\newblock Masked graph modeling for molecule generation.
\newblock {\em Nature communications}, 2021.

\bibitem[\protect\citeauthoryear{Meanwell}{2011}]{meanwell2011synopsis}
Nicholas~A Meanwell.
\newblock Synopsis of some recent tactical application of bioisosteres in drug
  design.
\newblock {\em Journal of medicinal chemistry}, 2011.

\bibitem[\protect\citeauthoryear{Mesquita \bgroup \em et al.\egroup
  }{2020}]{rethinkpooling2020}
D.~Mesquita, A.~H. Souza, and S.~Kaski.
\newblock Rethinking pooling in graph neural networks.
\newblock In {\em NeurIPS}, 2020.

\bibitem[\protect\citeauthoryear{Mikolov \bgroup \em et al.\egroup
  }{2013}]{mikolov2013distributed}
Tomas Mikolov, Ilya Sutskever, and others.
\newblock Distributed representations of words and phrases and their
  compositionality.
\newblock {\em NeurIPS}, 2013.

\bibitem[\protect\citeauthoryear{Narayanan \bgroup \em et al.\egroup
  }{2016}]{narayanan2016subgraph2vec}
Annamalai Narayanan, Mahinthan Chandramohan, and others.
\newblock subgraph2vec: Learning distributed representations of rooted
  sub-graphs from large graphs.
\newblock {\em arXiv:1606.08928}, 2016.

\bibitem[\protect\citeauthoryear{Pan \bgroup \em et al.\egroup
  }{2018}]{pan2018adversarially}
S.~Pan, R.~Hu, and others.
\newblock Adversarially regularized graph autoencoder for graph embedding.
\newblock In {\em IJCAI}, 2018.

\bibitem[\protect\citeauthoryear{Peng \bgroup \em et al.\egroup
  }{2020}]{peng2020graph}
Z.~Peng, W.~Huang, and others.
\newblock Graph representation learning via graphical mutual information
  maximization.
\newblock In {\em WWW}, 2020.

\bibitem[\protect\citeauthoryear{Perozzi \bgroup \em et al.\egroup
  }{2014}]{perozzi2014deepwalk}
Bryan Perozzi, Rami Al-Rfou, and Steven Skiena.
\newblock Deepwalk: Online learning of social representations.
\newblock In {\em KDD}, 2014.

\bibitem[\protect\citeauthoryear{Qiu \bgroup \em et al.\egroup
  }{2020}]{qiu2020gcc}
Jiezhong Qiu, Qibin Chen, and others.
\newblock Gcc: Graph contrastive coding for graph neural network pre-training.
\newblock In {\em KDD}, 2020.

\bibitem[\protect\citeauthoryear{Rong \bgroup \em et al.\egroup
  }{2020}]{rong2020self}
Yu~Rong, Yatao Bian, and others.
\newblock Self-supervised graph transformer on large-scale molecular data.
\newblock {\em NeurIPS}, 2020.

\bibitem[\protect\citeauthoryear{Sch\"{u}tt \bgroup \em et al.\egroup
  }{2017}]{NIPS2017_303ed4c6}
K.~Sch\"{u}tt, P.~Kindermans, et~al.
\newblock Schnet: A continuous-filter convolutional neural network for modeling
  quantum interactions.
\newblock In {\em NIPS}, 2017.

\bibitem[\protect\citeauthoryear{St{\"a}rk \bgroup \em et al.\egroup
  }{2021}]{stark20213d}
H.~St{\"a}rk, D.~Beaini, and others.
\newblock 3d infomax improves gnns for molecular property prediction.
\newblock {\em arXiv:2110.04126}, 2021.

\bibitem[\protect\citeauthoryear{Sun \bgroup \em et al.\egroup
  }{2021}]{sun2021mocl}
Mengying Sun, Jing Xing, and others.
\newblock Mocl: Contrastive learning on molecular graphs with multi-level
  domain knowledge.
\newblock {\em KDD}, 2021.

\bibitem[\protect\citeauthoryear{Sun}{2020}]{Sun2020InfoGraph:}
Infograph: Unsupervised and semi-supervised graph-level representation learning
  via mutual information maximization.
\newblock In {\em ICLR}, 2020.

\bibitem[\protect\citeauthoryear{Suresh \bgroup \em et al.\egroup
  }{2021}]{suresh2021adversarial}
Susheel Suresh, Pan Li, and others.
\newblock Adversarial graph augmentation to improve graph contrastive learning.
\newblock In {\em NeurIPS}, 2021.

\bibitem[\protect\citeauthoryear{T. \bgroup \em et al.\egroup
  }{2021}]{thakoor2021bootstrapped}
Shantanu T., Corentin T., and others.
\newblock Bootstrapped representation learning on graphs.
\newblock In {\em ICLR Workshop}, 2021.

\bibitem[\protect\citeauthoryear{Tang \bgroup \em et al.\egroup
  }{2015a}]{tang2015pte}
J.~Tang, M.~Qu, and Q.~Mei.
\newblock Pte: Predictive text embedding through large-scale heterogeneous text
  networks.
\newblock In {\em KDD}, 2015.

\bibitem[\protect\citeauthoryear{Tang \bgroup \em et al.\egroup
  }{2015b}]{tang2015line}
Jian Tang, Meng Qu, and others.
\newblock Line: Large-scale information network embedding.
\newblock In {\em WWW}, 2015.

\bibitem[\protect\citeauthoryear{Vaswani \bgroup \em et al.\egroup
  }{2017}]{vaswani2017attention}
Ashish Vaswani, Noam Shazeer, and others.
\newblock Attention is all you need.
\newblock {\em NIPS}, 2017.

\bibitem[\protect\citeauthoryear{Velickovic \bgroup \em et al.\egroup
  }{2019}]{velickovic2019deep}
P.~Velickovic, W.~Fedus, and others.
\newblock Deep graph infomax.
\newblock {\em ICLR}, 2019.

\bibitem[\protect\citeauthoryear{W. \bgroup \em et al.\egroup }{2021}]{heco}
Xiao W., Nian L., and others.
\newblock Self-supervised heterogeneous graph neural network with
  co-contrastive learning.
\newblock In {\em KDD}, 2021.

\bibitem[\protect\citeauthoryear{Wang \bgroup \em et al.\egroup
  }{2017}]{wang2017mgae}
C.~Wang, S.~Pan, and others.
\newblock Mgae: Marginalized graph autoencoder for graph clustering.
\newblock In {\em CIKM}, 2017.

\bibitem[\protect\citeauthoryear{Wang \bgroup \em et al.\egroup
  }{2019}]{wang2019heterogeneous}
Xiao Wang, Houye Ji, and others.
\newblock Heterogeneous graph attention network.
\newblock In {\em WWW}, 2019.

\bibitem[\protect\citeauthoryear{Wang \bgroup \em et al.\egroup
  }{2021a}]{wang2021curriculum}
Hui Wang, Kun Zhou, and others.
\newblock Curriculum pre-training heterogeneous subgraph transformer for top-$
  n $ recommendation.
\newblock {\em arXiv:2106.06722}, 2021.

\bibitem[\protect\citeauthoryear{Wang \bgroup \em et al.\egroup
  }{2021b}]{Wang2021MolCLRMC}
Y.~Wang, J.~Wang, and others.
\newblock Molclr: Molecular contrastive learning of representations via graph
  neural networks.
\newblock {\em ArXiv}, abs/2102.10056, 2021.

\bibitem[\protect\citeauthoryear{Wu \bgroup \em et al.\egroup
  }{2020}]{wu2020comprehensive}
Z.~Wu, S.~Pan, and others.
\newblock A comprehensive survey on graph neural networks.
\newblock {\em TNNLS}, 2020.

\bibitem[\protect\citeauthoryear{Xia and Ku}{2021}]{xia2021geometric}
T.~Xia and W.~Ku.
\newblock Geometric graph representation learning on protein structure
  prediction.
\newblock In {\em KDD}, 2021.

\bibitem[\protect\citeauthoryear{Xia \bgroup \em et al.\egroup
  }{2021a}]{xia2021towards}
Jun Xia, Haitao Lin, Yongjie Xu, Lirong Wu, Zhangyang Gao, Siyuan Li, and
  Stan~Z. Li.
\newblock Towards robust graph neural networks against label noise, 2021.

\bibitem[\protect\citeauthoryear{Xia \bgroup \em et al.\egroup
  }{2021b}]{xia2021debiased}
Jun Xia, Lirong Wu, Jintao Chen, Ge~Wang, and Stan~Z Li.
\newblock Debiased graph contrastive learning.
\newblock {\em arXiv:2110.02027}, 2021.

\bibitem[\protect\citeauthoryear{Xia \bgroup \em et al.\egroup
  }{2022a}]{xia2022simgrace}
Jun Xia, Lirong Wu, , Jintao Chen, Bozhen Hu, and Stan~Z. Li.
\newblock {SimGRACE: A Simple Framework for Graph Contrastive Learning without
  Data Augmentation}.
\newblock In {\em WWW}, 2022.

\bibitem[\protect\citeauthoryear{Xia \bgroup \em et al.\egroup
  }{2022b}]{Xia2022.02.03.479055}
Jun Xia, Jiangbin Zheng, Cheng Tan, Ge~Wang, and Stan~Z Li.
\newblock Towards effective and generalizable fine-tuning for pre-trained
  molecular graph models.
\newblock {\em bioRxiv}, 2022.

\bibitem[\protect\citeauthoryear{Xie \bgroup \em et al.\egroup
  }{2021}]{xie2021self}
Yaochen Xie, Zhao Xu, and others.
\newblock Self-supervised learning of graph neural networks: A unified review.
\newblock {\em arXiv:2102.10757}, 2021.

\bibitem[\protect\citeauthoryear{Xu \bgroup \em et al.\egroup
  }{2019}]{xu2018how}
Keyulu Xu, Weihua Hu, and others.
\newblock How powerful are graph neural networks?
\newblock In {\em ICLR}, 2019.

\bibitem[\protect\citeauthoryear{Xu \bgroup \em et al.\egroup
  }{2021}]{xu2021self}
M.~Xu, H.~Wang, and others.
\newblock Self-supervised graph-level representation learning with local and
  global structure.
\newblock In {\em ICML}, 2021.

\bibitem[\protect\citeauthoryear{Yang \bgroup \em et al.\egroup
  }{2020}]{textbrewer-acl2020-demo}
Z.~Yang, Y.~Cui, et~al.
\newblock {T}ext{B}rewer: {A}n {O}pen-{S}ource {K}nowledge {D}istillation
  {T}oolkit for {N}atural {L}anguage {P}rocessing.
\newblock In {\em ACL: System Demonstrations}, 2020.

\bibitem[\protect\citeauthoryear{You \bgroup \em et al.\egroup
  }{2020}]{You2020GraphCL}
Y.~You, T.~Chen, and others.
\newblock Graph contrastive learning with augmentations.
\newblock In {\em NeurIPS}, 2020.

\bibitem[\protect\citeauthoryear{You \bgroup \em et al.\egroup
  }{2021}]{you2021graph}
Y.~You, T.~Chen, and others.
\newblock Graph contrastive learning automated.
\newblock {\em ICML}, 2021.

\bibitem[\protect\citeauthoryear{You \bgroup \em et al.\egroup
  }{2022}]{you2022bringing}
Y.~You, T.~Chen, Z.~Wang, and Y.~Shen.
\newblock Bringing your own view: Graph contrastive learning without
  prefabricated data augmentations.
\newblock In {\em WSDM}, 2022.

\bibitem[\protect\citeauthoryear{Zhang \bgroup \em et al.\egroup
  }{2020}]{zhang2020graph}
J.~Zhang, H.~Zhang, C.~Xia, and L.~Sun.
\newblock Graph-bert: Only attention is needed for learning graph
  representations.
\newblock {\em arXiv preprint arXiv:2001.05140}, 2020.

\bibitem[\protect\citeauthoryear{Zhang \bgroup \em et al.\egroup
  }{2021a}]{zhang2021canonical}
H.~Zhang, Q.~Wu, and others.
\newblock From canonical correlation analysis to self-supervised graph neural
  networks.
\newblock In {\em NeurIPS}, 2021.

\bibitem[\protect\citeauthoryear{Zhang \bgroup \em et al.\egroup
  }{2021b}]{zhang2021motif}
Zaixi Zhang, Qi~Liu, and others.
\newblock Motif-based graph self-supervised learning for molecular property
  prediction.
\newblock {\em NeurIPS}, 2021.

\bibitem[\protect\citeauthoryear{Zhu \bgroup \em et al.\egroup
  }{2020}]{Zhu:2020vf}
Yanqiao Zhu, Yichen Xu, Feng Yu, Qiang Liu, Shu Wu, and Liang Wang.
\newblock {Deep Graph Contrastive Representation Learning}.
\newblock In {\em ICML Workshop}, 2020.

\bibitem[\protect\citeauthoryear{Zhu \bgroup \em et al.\egroup
  }{2021a}]{Zhu2021DualviewMP}
Jinhua Zhu, Yingce Xia, Tao Qin, Wen gang Zhou, Houqiang Li, and Tie-Yan Liu.
\newblock Dual-view molecule pre-training.
\newblock {\em ArXiv}, abs/2106.10234, 2021.

\bibitem[\protect\citeauthoryear{Zhu \bgroup \em et al.\egroup
  }{2021b}]{zhu2021an}
Yanqiao Zhu, Yichen Xu, Qiang Liu, and Shu Wu.
\newblock An empirical study of graph contrastive learning.
\newblock In {\em NeurIPS Datasets and Benchmarks Track}, 2021.

\bibitem[\protect\citeauthoryear{Zhu \bgroup \em et al.\egroup
  }{2021c}]{Zhu:2021wh}
Yanqiao Zhu, Yichen Xu, Feng Yu, Qiang Liu, Shu Wu, and Liang Wang.
\newblock {Graph Contrastive Learning with Adaptive Augmentation}.
\newblock In {\em WWW}, 2021.

\bibitem[\protect\citeauthoryear{Zhu \bgroup \em et al.\egroup
  }{2022}]{Zhu:2020ui}
Yanqiao Zhu, Yichen Xu, Hejie Cui, Carl Yang, Qiang Liu, and Shu Wu.
\newblock Structure-enhanced heterogeneous graph contrastive learning.
\newblock In {\em SDM}, 2022.

\end{thebibliography}

\end{document}